# Enhancing Deep Neural Network Saliency Visualizations with Gradual Extrapolation

**Tomasz Szandała**[1]

[1]Wroclaw University of Science and Technology, Wroclaw, Poland

Corresponding author: Tomasz Szandała (e-mail: Tomasz.Szandala@pwr.edu.pl).

**ABSTRACT** In this paper, an enhancement technique for the class activation mapping methods such as gradient-weighted class activation maps or excitation backpropagation is proposed to present the visual explanations of decisions from convolutional neural network-based models. The proposed idea, called Gradual Extrapolation, can supplement any method that generates a heatmap picture by sharpening the output. Instead of producing a coarse localization map that highlights the important predictive regions in the image, the proposed method outputs the specific shape that most contributes to the model output. Thus, the proposed method improves the accuracy of saliency maps. The effect has been achieved by the gradual propagation of the crude map obtained in the deep layer through all preceding layers with respect to their activations. In validation tests conducted on a selected set of images, the faithfulness, interpretability, and applicability of the method are evaluated. The proposed technique significantly improves the localization detection of the neural networks' attention at low additional computational costs. Furthermore, the proposed method is applicable to a variety deep neural network models.

The code for the method can be found at https://github.com/szandala/gradual-extrapolation

## I. INTRODUCTION

Deep convolutional neural networks (CNN) have significantly improved the performance of computer vision systems[1]. Initially developed for image classification, CNN-based methods have recently been employed in pixel-level image segmentation [2,3]. Although segmentation methods capture a large amount of information, they require expensive training data labeling. Therefore, classification networks[4] are still used when the problem cannot be formulated as a segmentation task or when the pixel-wise labeling is extremely expensive (as observed in many applications).

Unfortunately, deep networks provide no reasoning based on which part of the image contributes to the network's decision[5]. To reveal and visualize the factors contributing to the model's output, researchers have proposed various visualization techniques[6,7,8,9]. Among these techniques, class activation maps (CAMs) can efficiently demonstrate the discriminative features of the input image; however, the resulting saliency map is quite fuzzy.

Therefore, a simple yet efficient method is proposed that reduces the coarseness of the obtained heatmaps. Using a weighted mask multiplication technique, the sharpness of the features shown in the heatmap is improved herein. Although the proposed method can be combined with almost any saliency map for CNNs, herein, its use is demonstrated with gradient-weighted CAM (Grad-CAM) and Excitation Backpropagation.

## II. STATE-OF-THE-ART

Before CAM, the saliency map of an image was most commonly obtained via occlusion. A small part of the image was hidden, and the remainder of the image was classified using the model. Amore altered final score signified the greater importance of the hidden region for the classifier. However, occlusion methods are slow because image processing must be restarted for each occluded region. Moreover, they do not consider the correlations between areas[12].

The introduction of CAM encouraged the development of numerous other heatmap-generating methods that visualize CNN decisions. Nonetheless, their common result is a stain that marks the most significant area.

In the original CAM approach, a trained model is modified by discarding all fully connected layers at the top of the network and introducing a global average pooling layer, followed by a single fully connected layer[11]. This simplified architecture allows the CAM to generate a heatmap for each output class. Thus, it sums the activations of the last convolutional layer rescaled by the weights of the newly introduced fully connected layer associated with the selected class.

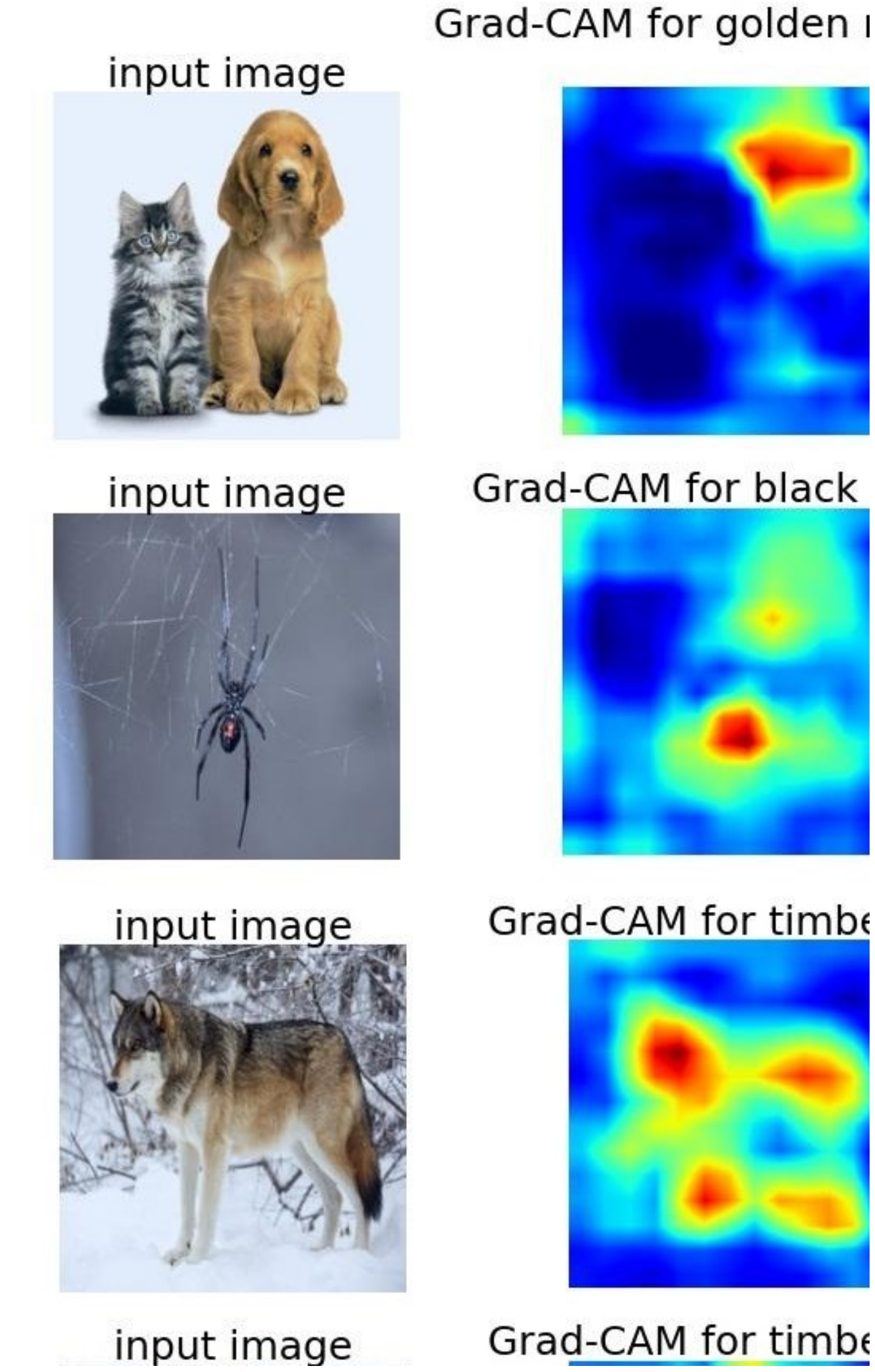

**Fig. 1.** Basic Grad-CAM results of the selected images

Grad-CAM generalizes the aforementioned CAM to any deep neural network architecture, extending its use to a wider range of networks while removing the need to modify the network architecture.

A variant of Grad-CAM called Grad-CAM++ computes the higher-order derivatives to increase the localization accuracy of Grad-CAM, particularly when the image contains multiple occurrences of the same objects. Smooth Grad-CAM++[4] adopts the smooth grad function[10] to smoothen the gradients obtained using Grad-CAM++. This variant provides visually sharpened heatmaps. The smooth grad method creates multiple noisy versions of the same input image, which can be regarded as a simple form of image augmentation.

Apart from smooth Grad-CAM++ (which adds noise to images), the interpretability tools based on heat maps do not perform augmentation processes. Moreover, none of the existing solutions address heatmap computation as a super-resolution problem.

The authors of excitation backpropagation proposed a different approach [8], which is intuitively simple but provides empirically effective explanations. Excitation Backpropagation ignores the nonlinearities in the backward pass in the network and generates heatmaps that preserve the evidence for or against a class predicted using the network. When the evidence against a certain class is subtracted from the evidence for that class, a visually sharp contrastive variant is obtained. The visualization accuracy of this approach is higher than those of previous CAMs (Figures 1 and 2). However, this approach highlights the important area rather than the object itself.

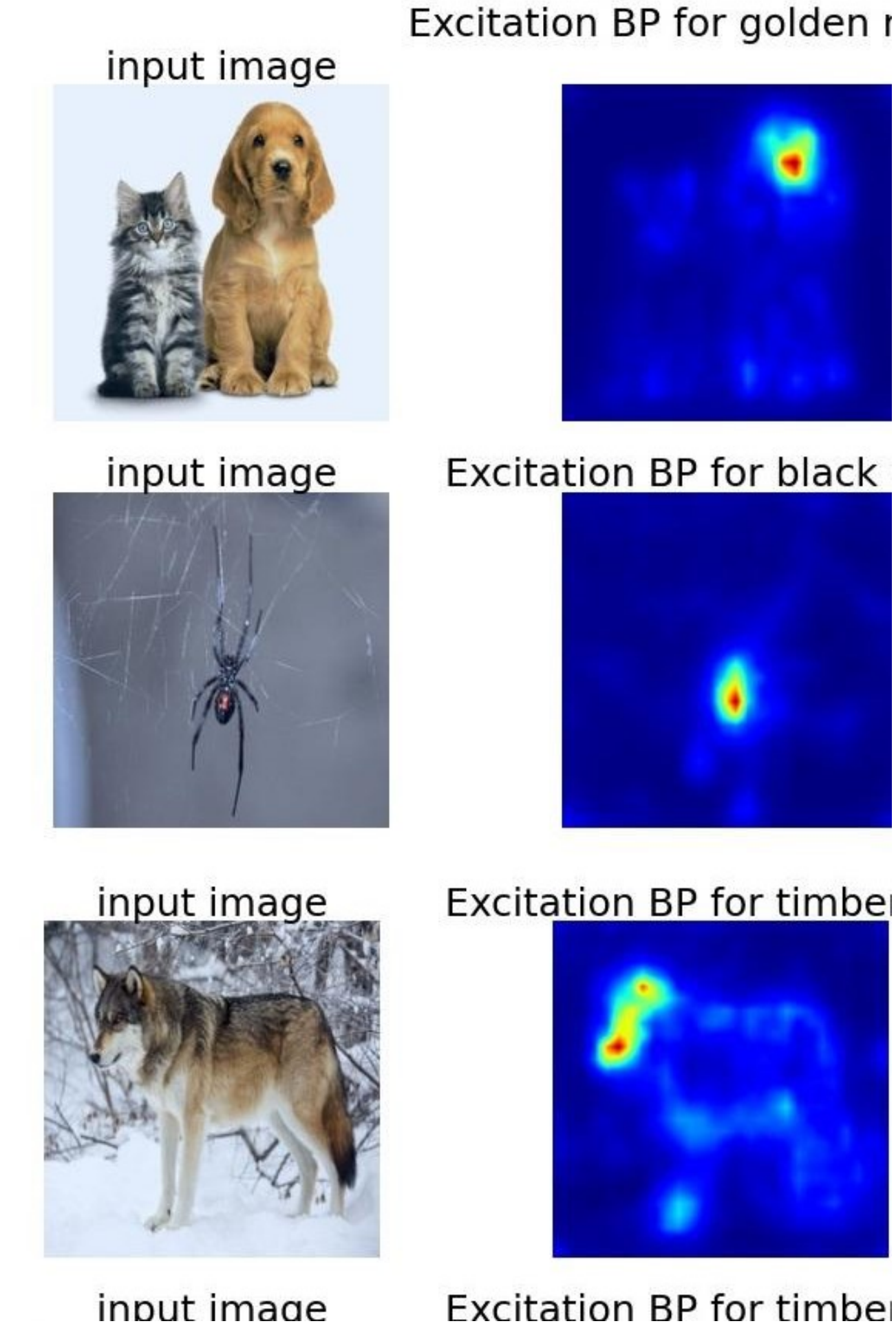

**Fig. 2.** Basic contrastive excitation backpropagation results of the selected images

Saliency methods aim to explain the predictions obtained using deep neural networks. These methods lack reliability when the explanation is insensitive to the object shape, which may be the factual factor contributing to the model prediction[16,17]. All the described methods focus on determining the area, often in the deep layer of the network. Therefore, projecting the deep map to the original image may considerably reduce the accuracy of the map; thus, the important trait of the classification reasoning may be omitted[20].

## III. GRADUAL EXTRAPOLATION

Gradual extrapolation is based on a simple concept. The map is expanded to the size of the preceding layer and then multiplied by the matrix representing the weights of the contributions from the given layer.

Here, the selection of the contribution matrix is important. A simple matrix of mean values (m) is selected, and all masks of the given layer are summed with respect to the channels. For instance, assume that there are $C$ channels, each with a mask of height $h$ and width $w$. Then, one matrix of dimensions $h \times w$ is obtained follows:

$$\forall_i \in h, \forall_j \in w : m_{i,j} = \left(\sum_{c=1}^{C} x_{i,j,c}\right) / C. \quad (1)$$

This process is repeated for each MaxPool[13] layer until the input dimensions are restored. The concept is superficially represented inFigure3.

In other words, for each preceding layer, a matrix of importance values is determined. Therefore, simply the mean values of the single cells across all channels in the selected layer are selected. However, this unsophisticated approach can be improved and refined with more advanced techniques.

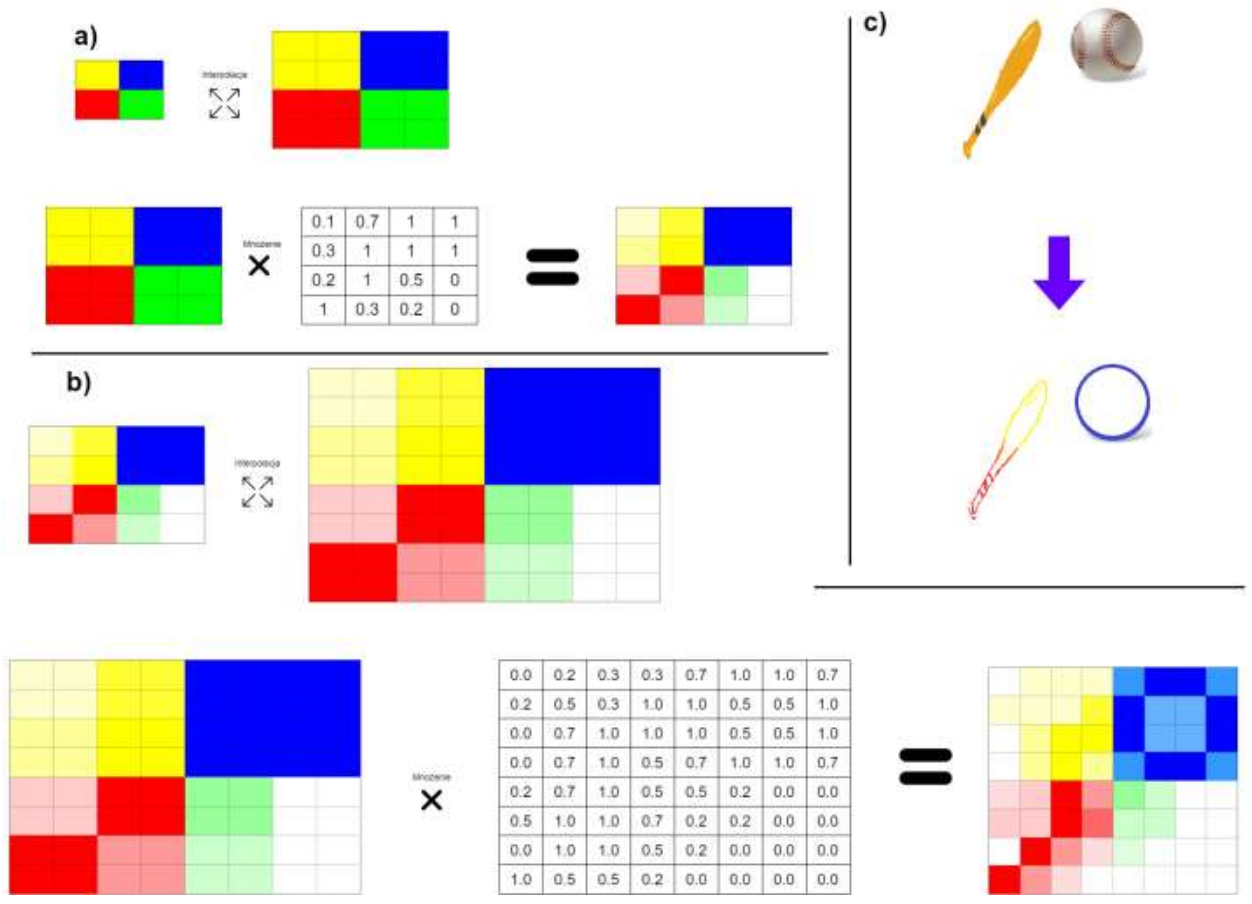

**Fig. 3.**Schematic concept of gradual extrapolation. Section a) and b) represents gradual extrapolation transition from an image of size 2 × 2 to 8 × 8. Section c) shows the concept of transition between image and its restoration using gradual extrapolation

To elucidate this process, the contribution matrix is divided by its maximal value, which restricts the possible values within the unit interval range. As the method can be applied to a model of any depth, restricting the values to 0 and 1 also prevents method overflow.

## IV. METHOD APPLICATION

First, the following question arises: are gradual Grad-CAM and gradual excitation backpropagation visualizations more class-discriminative than the previous techniques? In other words, can the end user trust the visualized map? To answer this question, the results of several classifications and visualizations are compared using the VGG-16 architecture network[14].

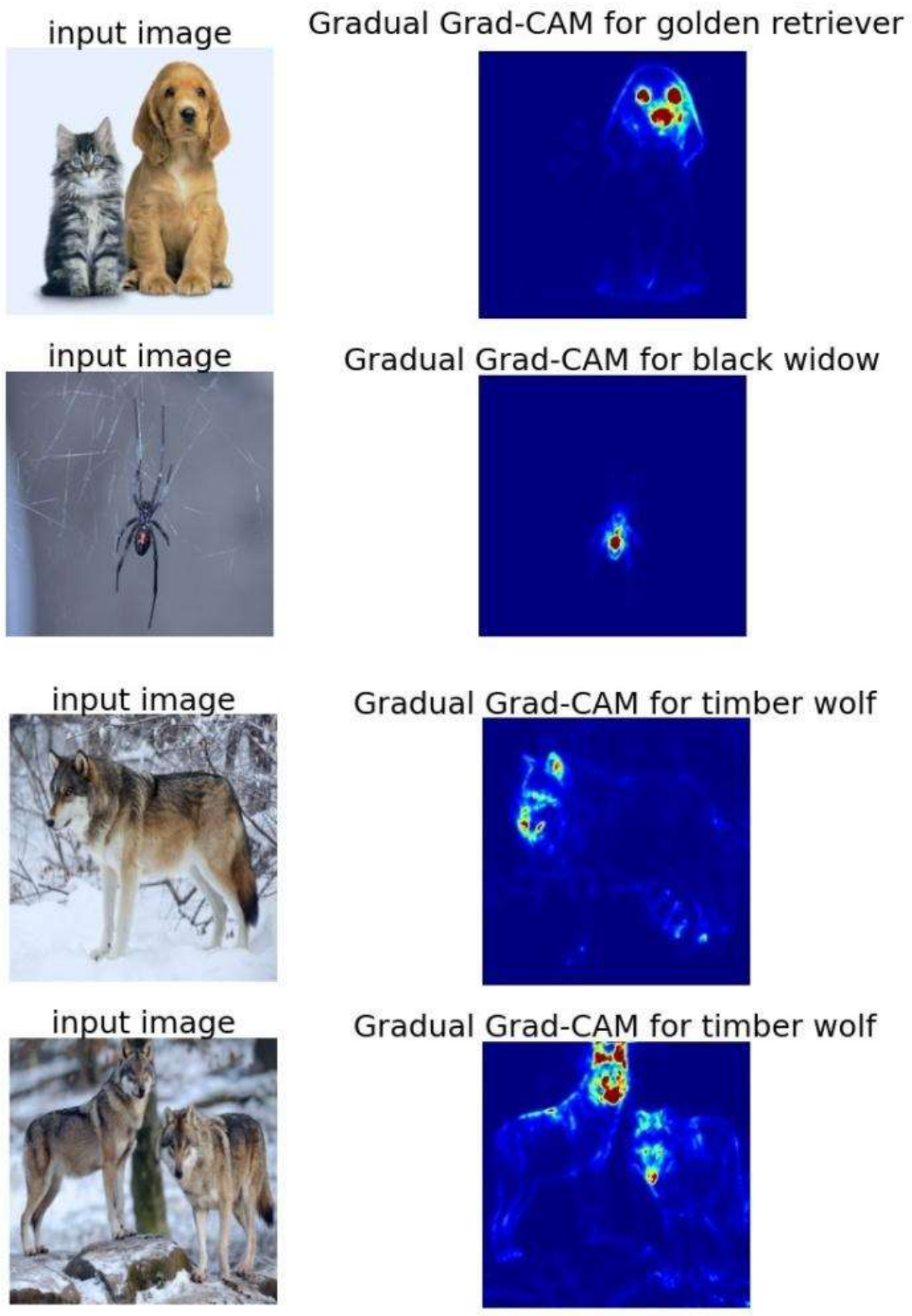

**Fig. 4.** Results of Grad-CAM enhanced using Gradual Extrapolation

The results of implementing gradual extrapolation in Grad-CAM and contrastive excitation backpropagation are presented in Figures 4 and 5, respectively. Gradual extrapolation enhances the shape of the object that most contributes to the classification. For example, the classification of the golden retriever puppy highlights the eyes and nose of the dog (first row in Figures 4 and 5). Apparently, the ears and the remainder of the body, which are strongly identifying features of humans, are ignored by the network.

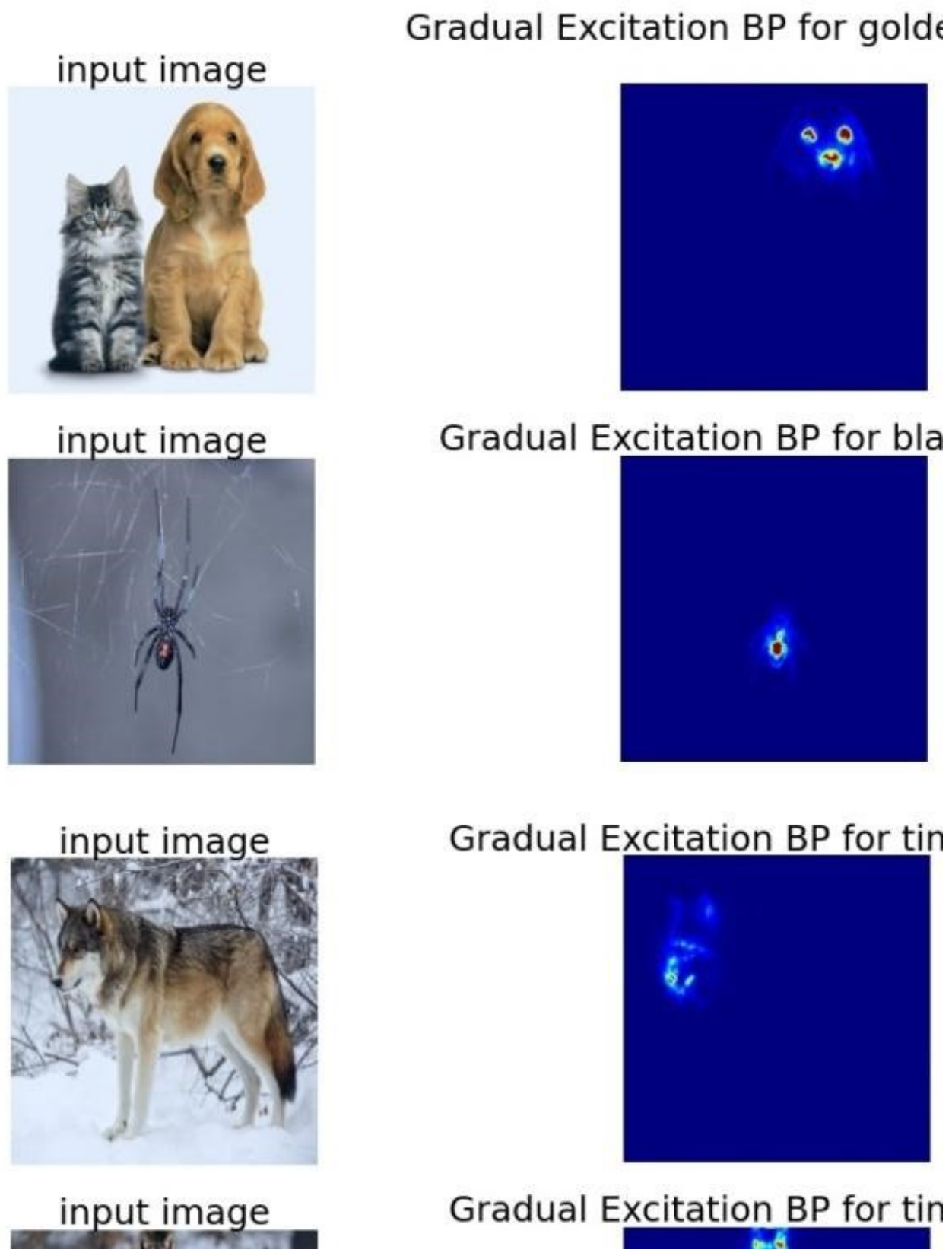

**Fig. 5.**Resultsof contrastive excitation backpropagation enhanced using gradual extrapolation

However, the classification of the black widow spider highlights the famous red mark on the spider's abdomen (second row in Figures 4 and 5).

Finally, the two images of wolves were investigated. The proposed improvement enhanced the results of the original techniques. The mouths of the wolves appear to dominate the picture classification. Interestingly, the basic Grad-CAM highlights the wolf bodies in the second wolf image. However, after Gradual Extrapolation, the most significant features are found to be around the head of the animal (bottom rows in Figures 1 and 4). These results align with those obtained using the Excitation Backpropagation approach, which highlights mainly the heads both before and after Gradual Extrapolation (bottom rows in Figures 2 and 5).

## V. METHOD EVALUATION

Objective assessment is crucial to quantify the degree of accuracy of an explanation. Unfortunately, evaluating explanations is difficult because it is generally impossible to achieve ground-truth explanations. Obtaining the ground-truth explanations would require a perfect expert to understand how the deep neural network formulates its answer, thereby creating a chicken–egg problem. Common models are usually evaluated based on the utility(expected risk) of their decision behavior (e.g., [27]).

Samek et al. proposed several methods to evaluate the models' explanations [15, 22], from them we have assembled a set of metrics that measure: faithfulness, interpretability and applicability (FIA).

### Faithfulness

The first criterion of an explanation is to reliably and comprehensively represent the local decision structure of the analyzed model. To assess such a property of the model, a proposed technique is "pixel flipping" [15]. The pixel-flipping procedure assesses whether removing the features highlighted by the explanation, as the most relevant, decreases the network prediction abilities.

Pixel flipping runs from the most to the least relevant input features, iteratively removing them and monitoring the evolution of the neural network output. The series of recorded decaying prediction scores can be plotted, where the faster the curve decreases, the more faithful the explanation method with respect to the neural network's decision. The pixel-flipping curve can be computed for a single example or averaged over an entire data set to achieve a global estimate of the faithfulness of an explanation algorithm under study.

For this assessment, random 1000 images are selected from the ImageNet dataset and classified using VGG-16, yielding 1.00 confidence score (based on the Softmax formula over the network). In Figure 6, Gradual Extrapolation offers noticeable improvements over conventional Grad-CAM and minor improvements over Excitation Backpropagation.

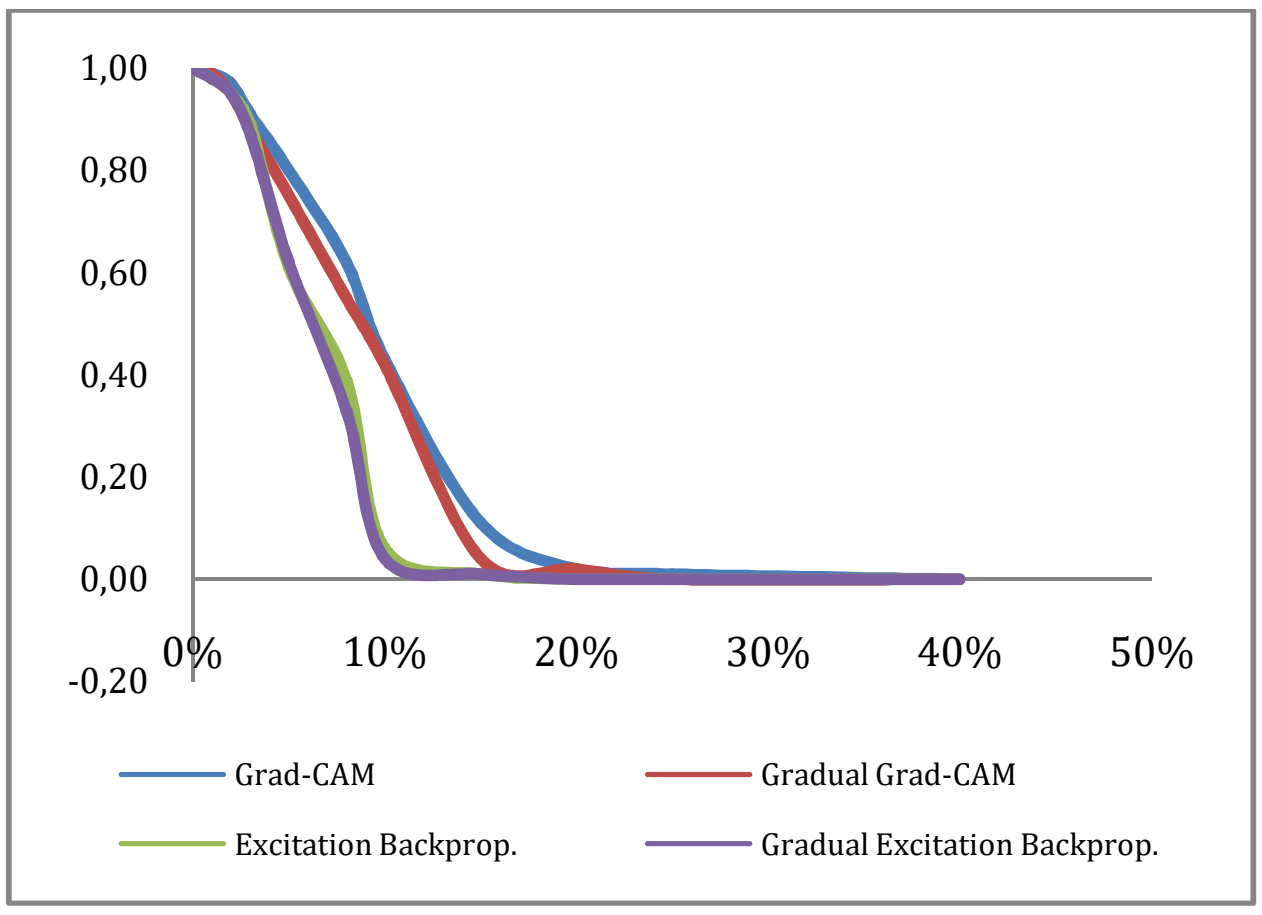

**Fig. 6.** Results of pixel-flipping impact on model's classification accuracy

### Interpretability

In our opinion the hardest to estimate, therefore we need to get back to the root of the explanation methods concept. The aim is to reduce the information from the original object and only retain the elements that play the highest role in classification[28]. Based on this consideration, a test is established that computes the number of pixels of the original image remaining after its truncation only in salient areas. This is can be considered as interpretability and is one of the metrics proposed by Hooker et al.[16].

**Table 1.** Significant regions and their sizes (in pixels) based on vanilla Grad-CAM(left) and Grad-CAM using Gradual Extrapolated (right)

| | |
|---|---|
| 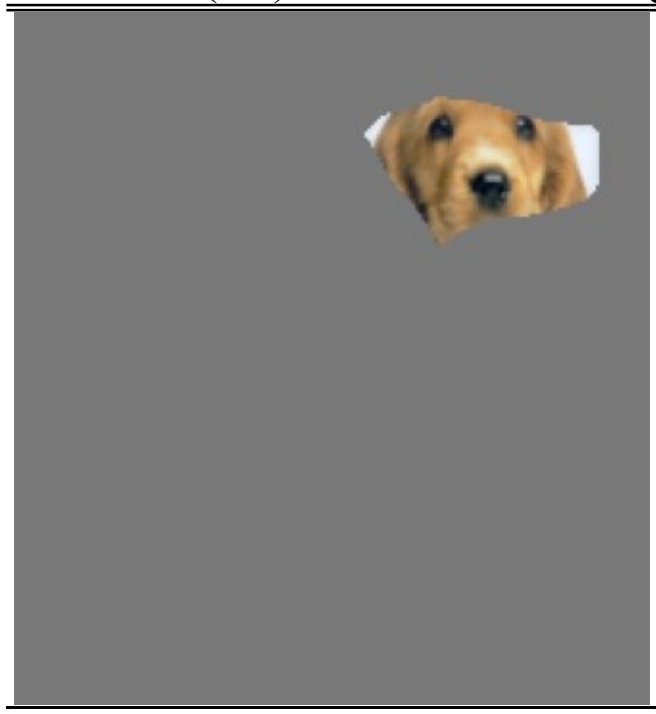 | 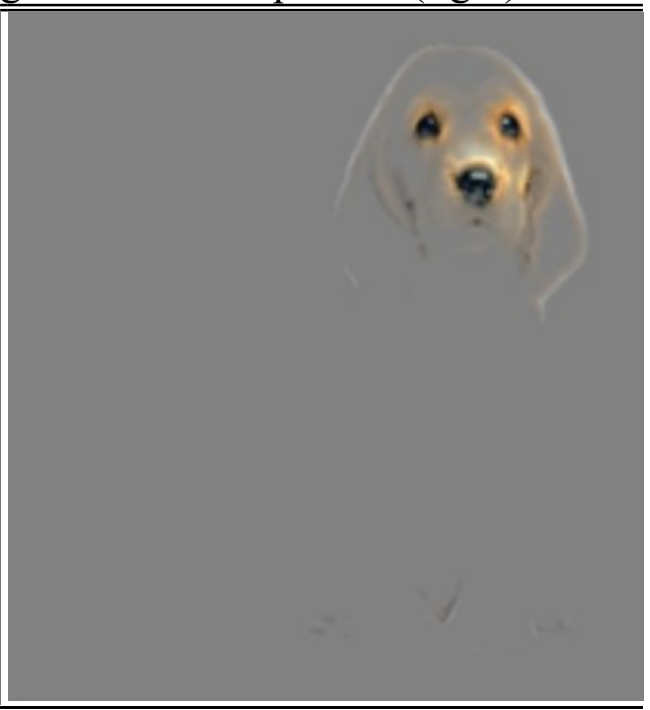 |
| Significant pixels: 4713 | Significant pixels: 3178 |
| 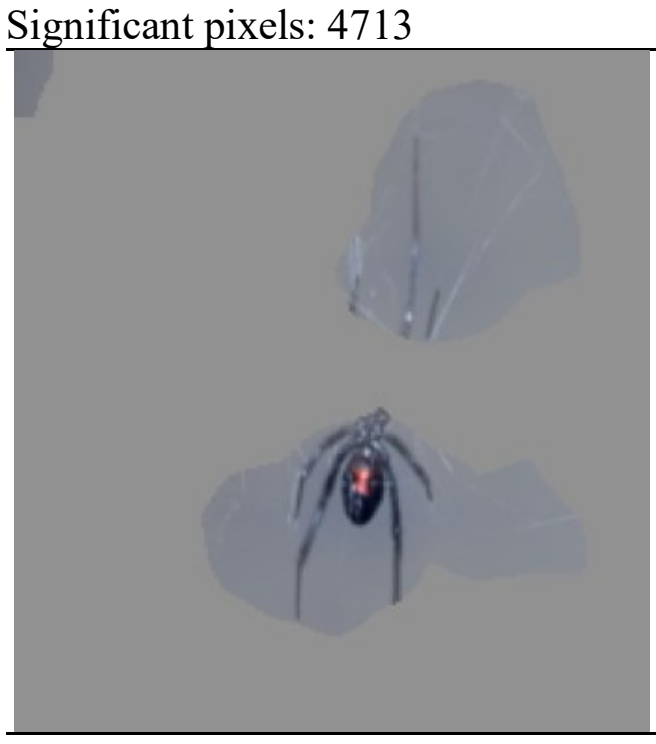 | 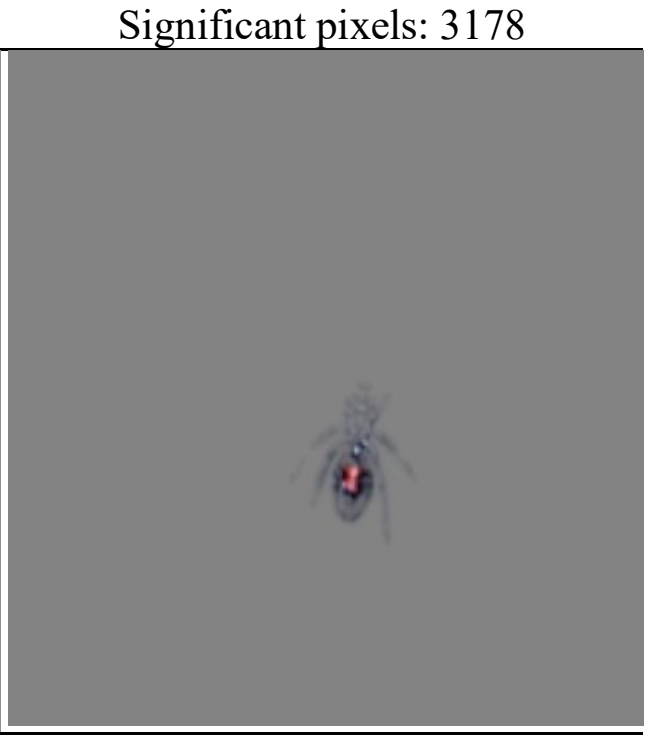 |
| Significant pixels: 16576 | Significant pixels: 1406 |
| 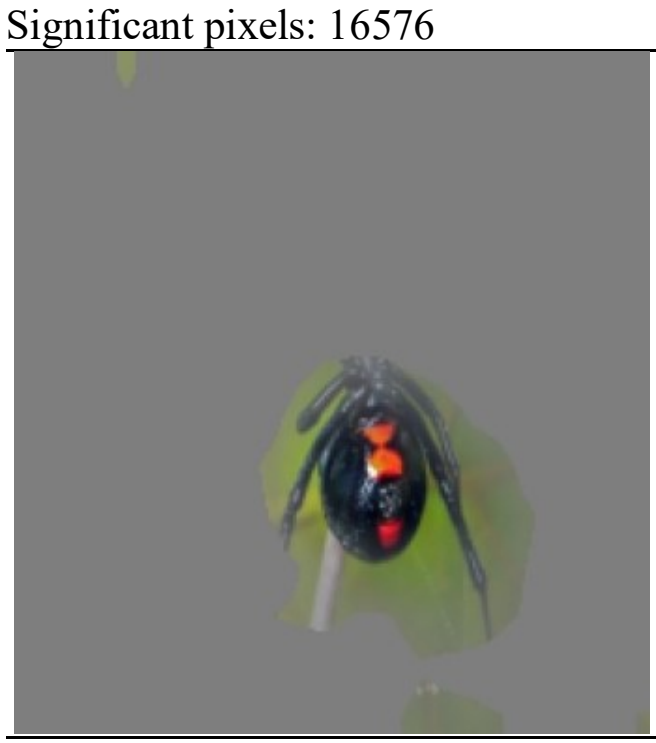 | 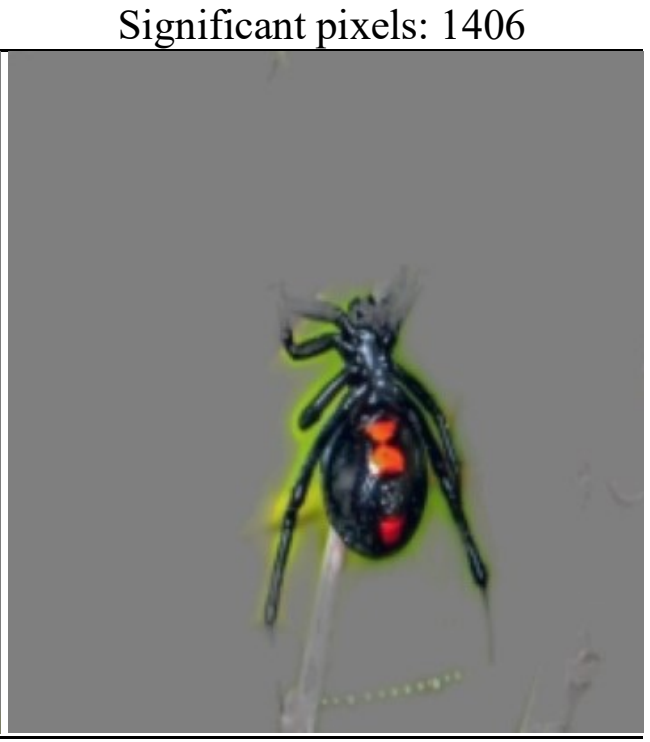 |
| Significant pixels: 9528 | Significant pixels: 10039 |
| 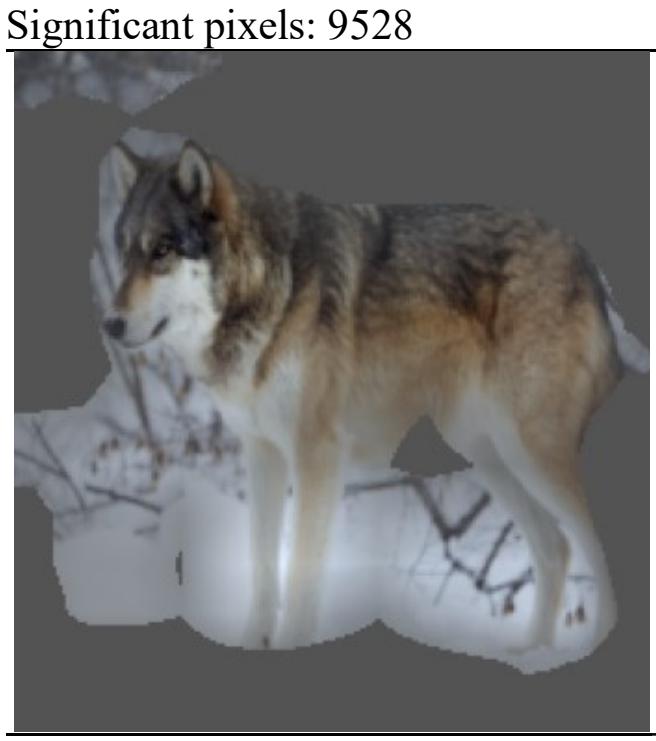 | 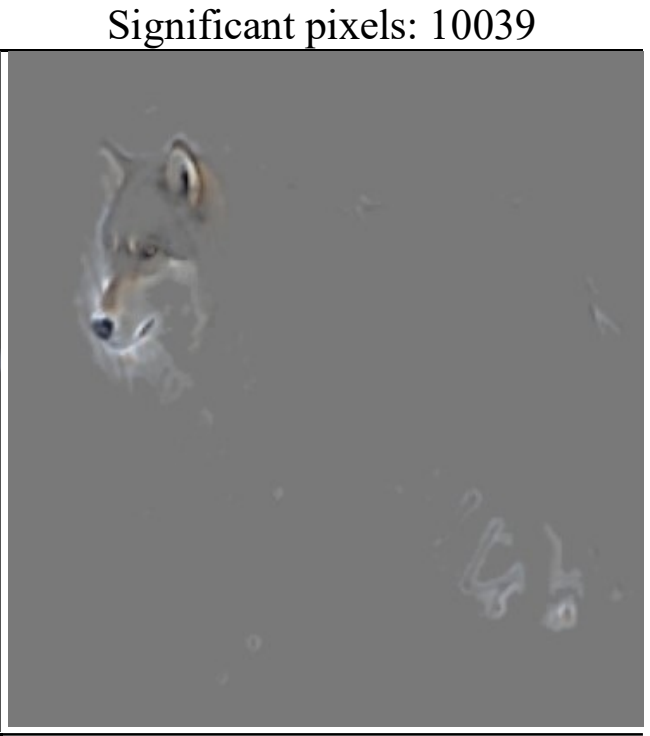 |
| Significant pixels: 29527 | Significant pixels: 4452 |
| 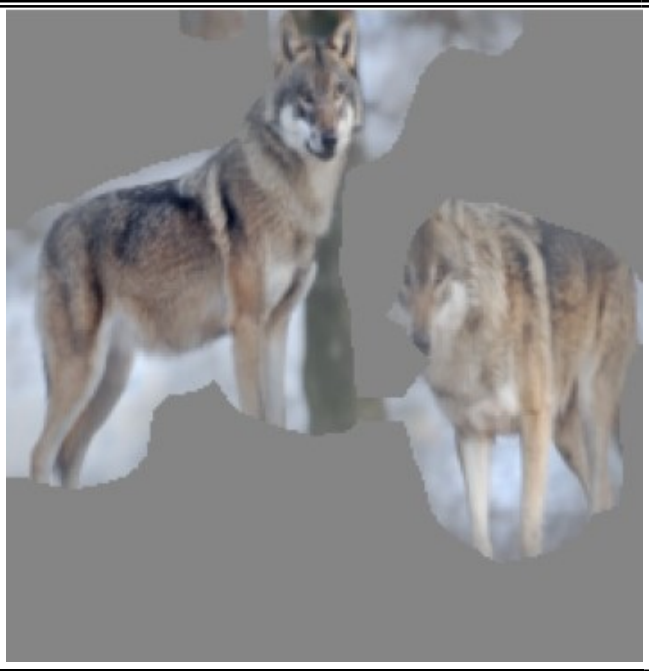 | 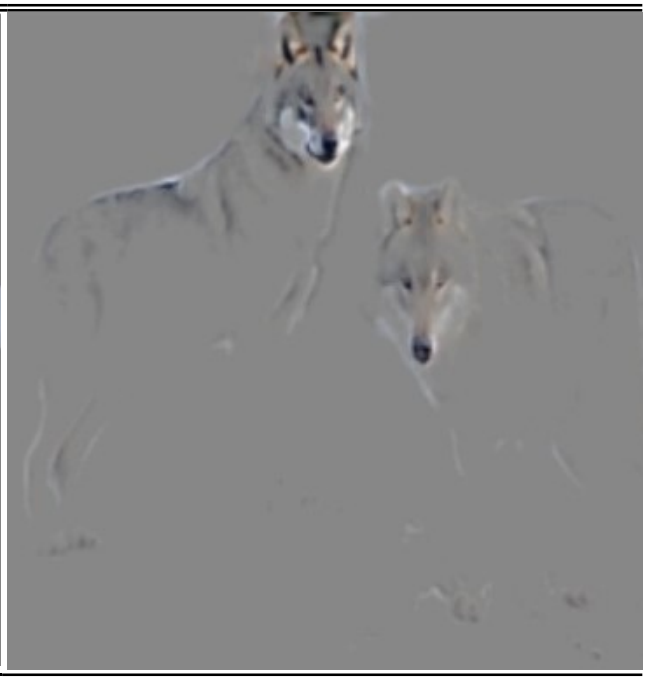 |
| Significant pixels: 22673 | Significant pixels: 5178 |

Tables 1 and 2 show a comparison of image regions highlighted as important using the selected methods. Generally, a firm decrease in the size of salient areas between the conventional methods and their enhanced versions can be observed. The only exception here is one of the black widows. The number of important pixels is higher. However, the enhanced version is more focused on animals, which can still be improved from a human perspective.

Moreover, if the black widow is less prominent, the improvement will be noticeable. The Grad-CAM's setback is the splitting into three main areas that contribute to the classification. Alternatively, gradual extrapolation is limited to only one object, which significantly better corresponds to the classification.

**Table 2.** Significant regions and their sizes (in pixels) based on Contrastive Excitation Backpropagation(left) and Gradual Extrapolated version(right)

| | |
|---|---|
| 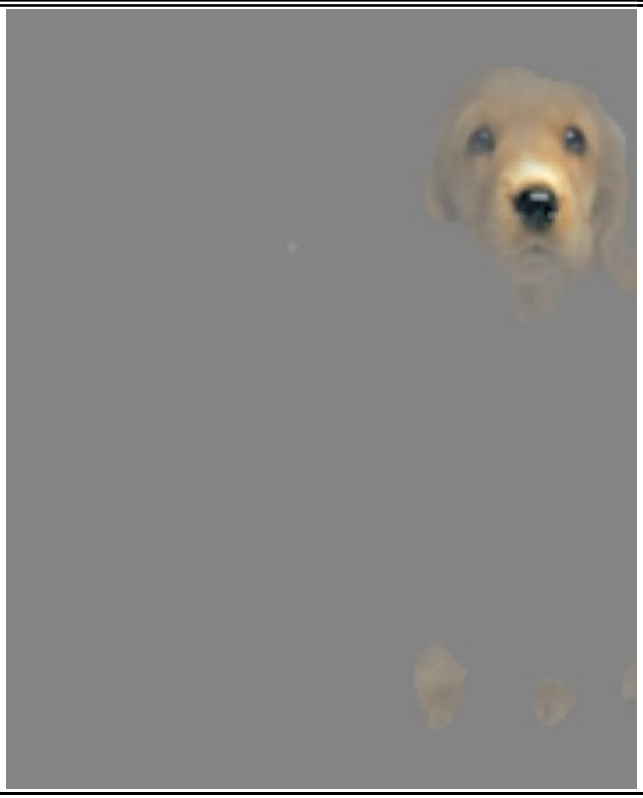 | 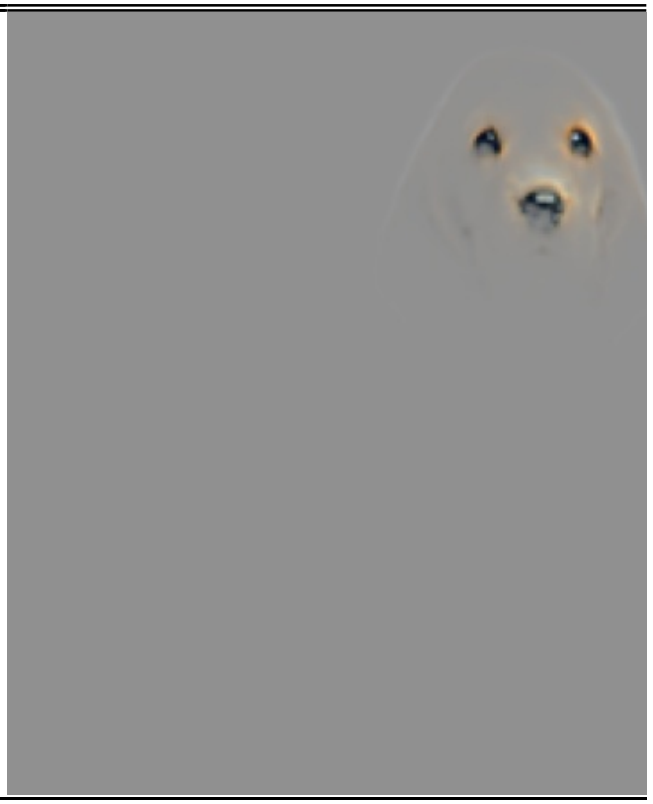 |
| Significant pixels: 5557 | Significant pixels: 3259 |
| 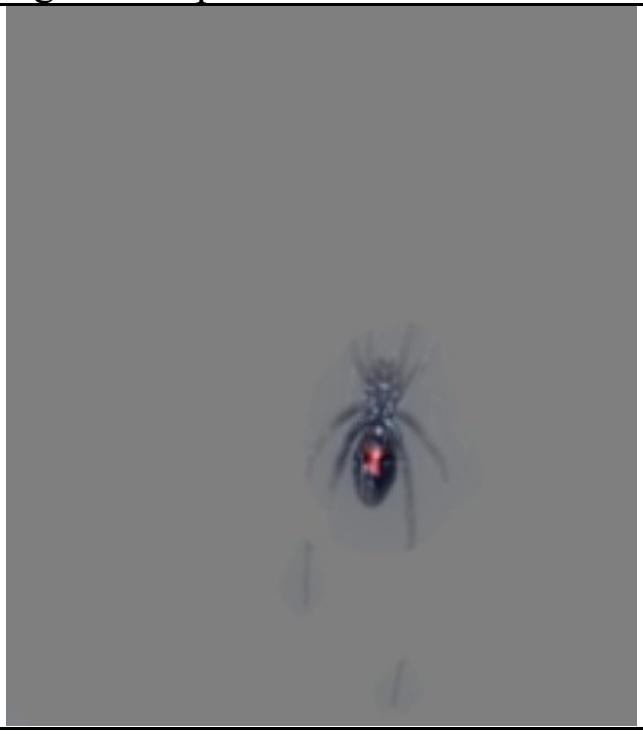 | 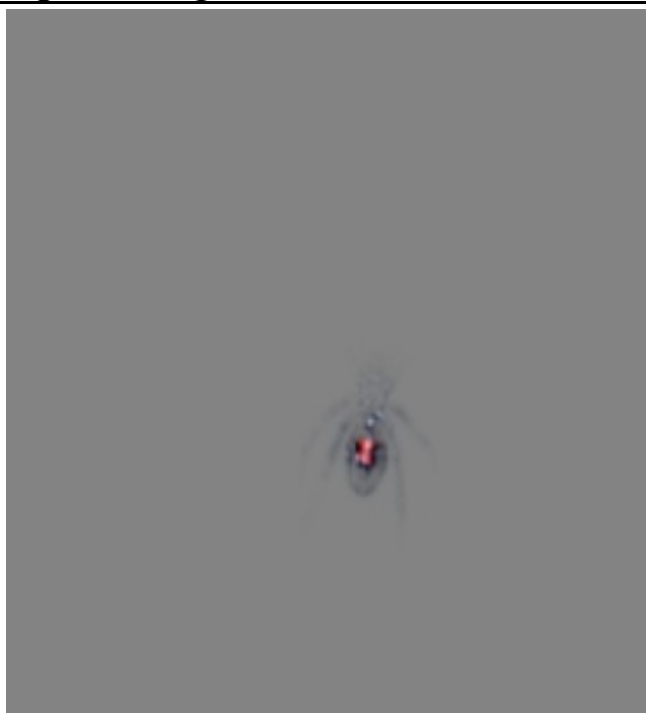 |
| Significant pixels: 8162 | Significant pixels: 1341 |

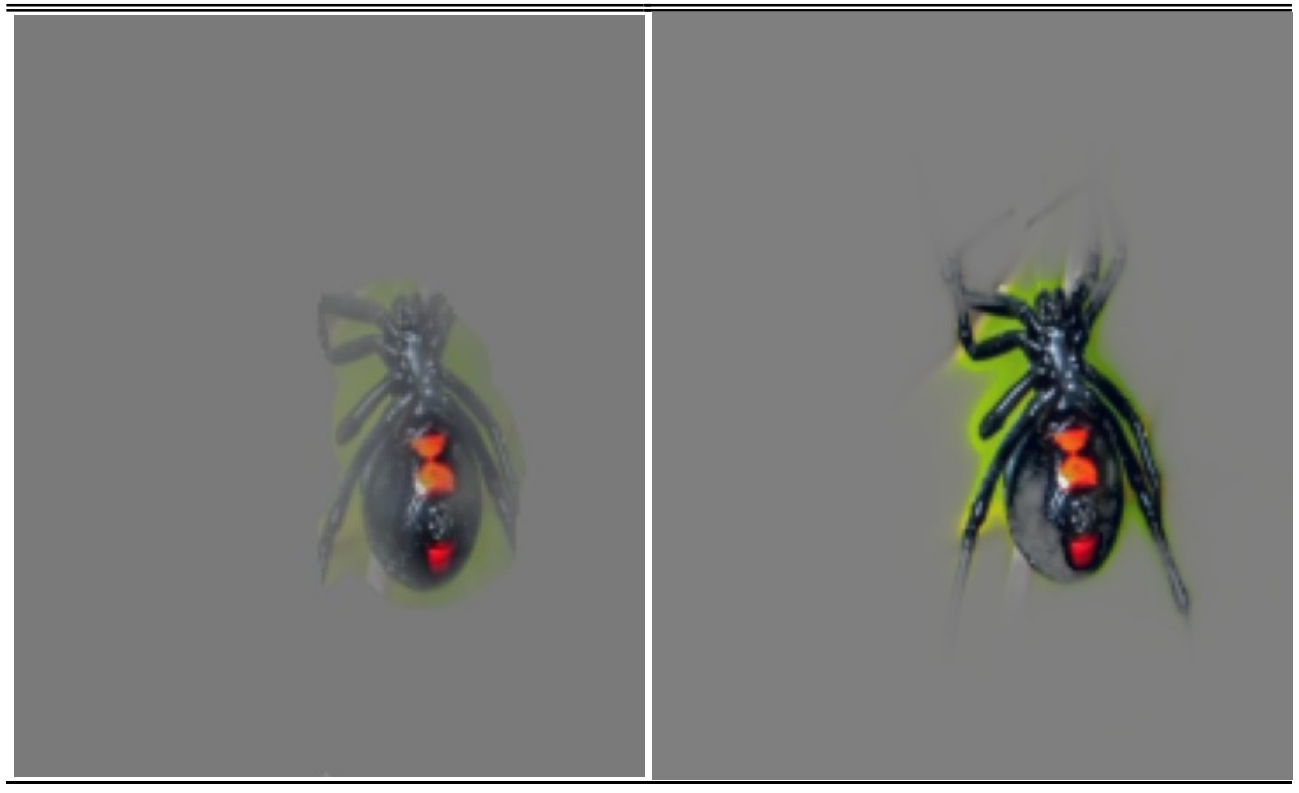

Significant pixels: 4749 Significant pixels: 8835

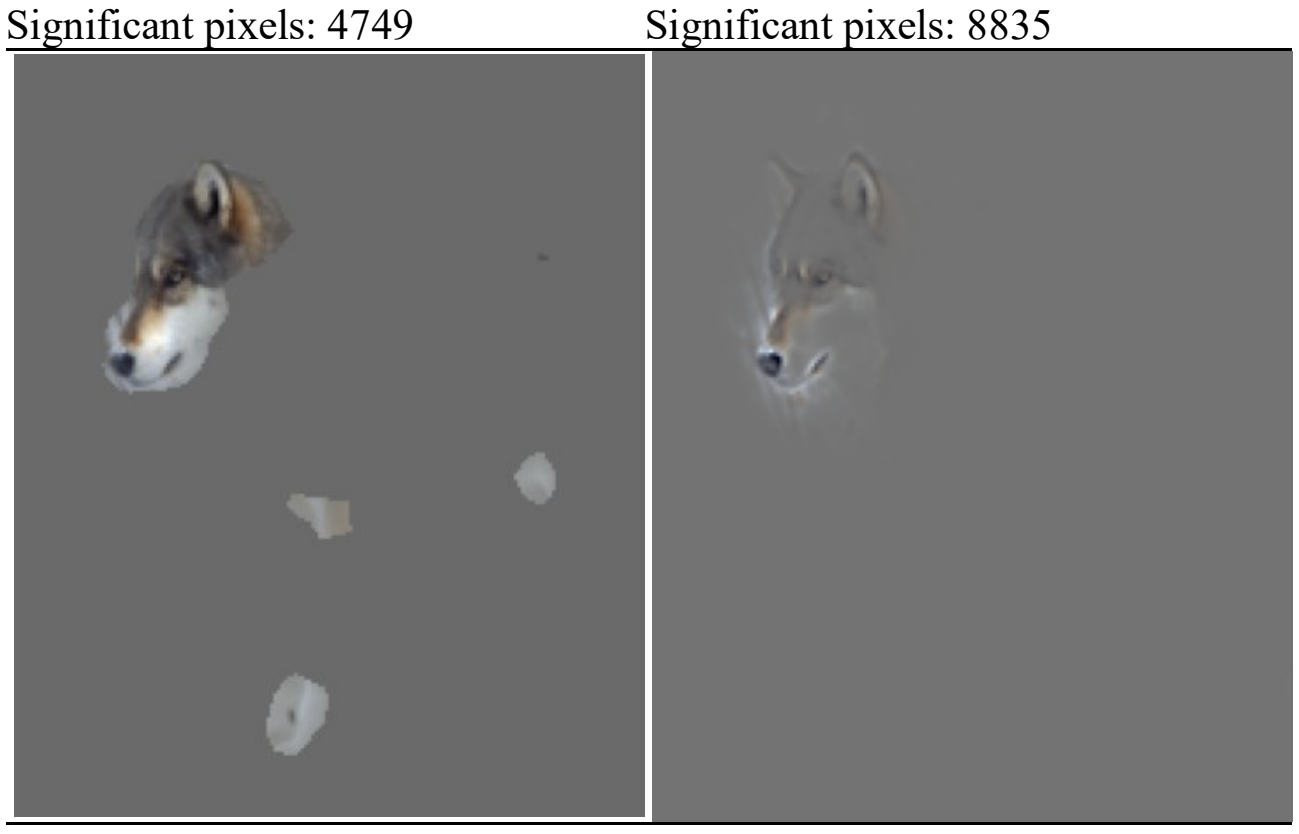

Significant pixels: 9994 Significant pixels: 3983

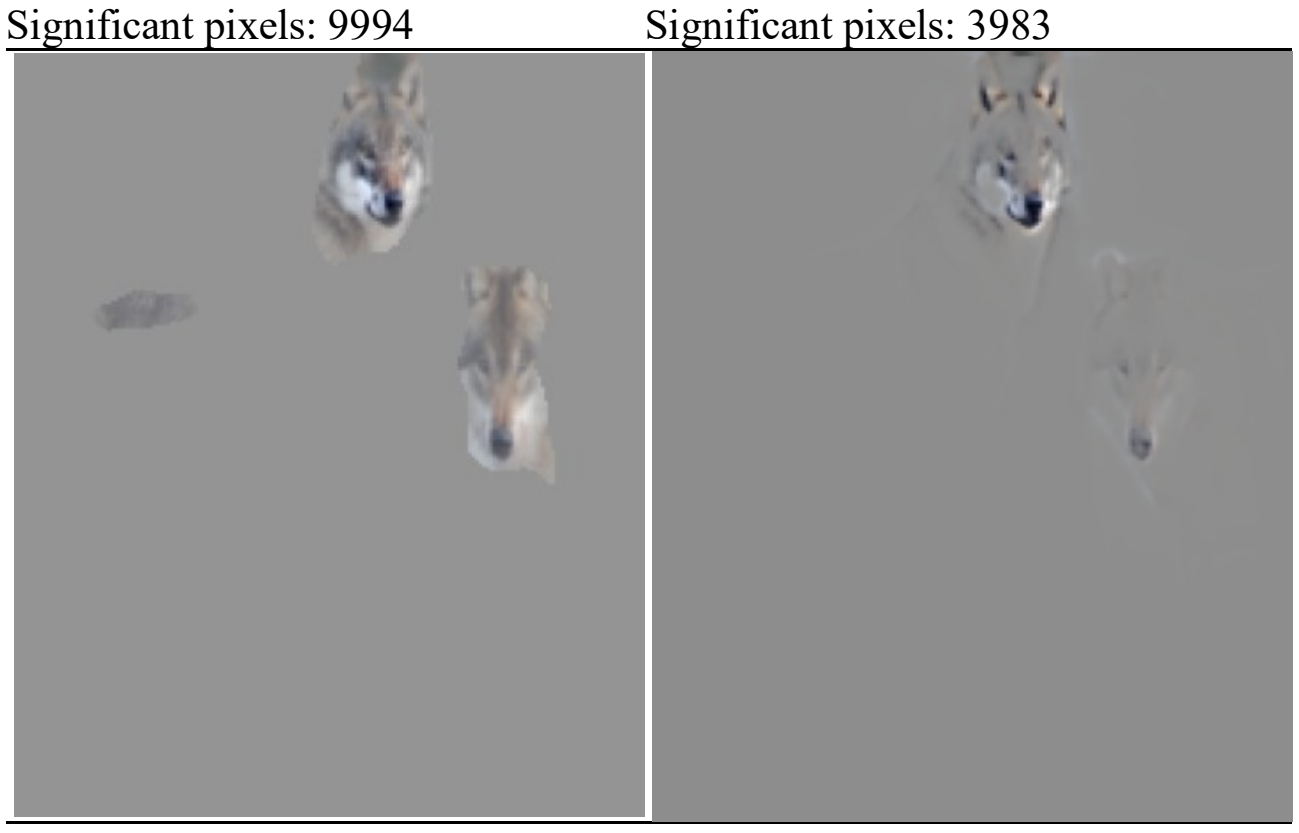

Significant pixels: 11855 Significant pixels: 7883

## Applicability

Faithfulness and interpretability do not completely determine the overall usefulness of the explanation method. To characterize the usefulness, it is also necessary to determine whether the explanation method can be implemented in various models, at least to most state-of-the-art models, and whether the explanation can be obtained quickly enough with finite computational resources.

To measure applicability, the proposed method is applied to several different models and their computational time is measured. The results related to the computational time are shown in Figure 7.

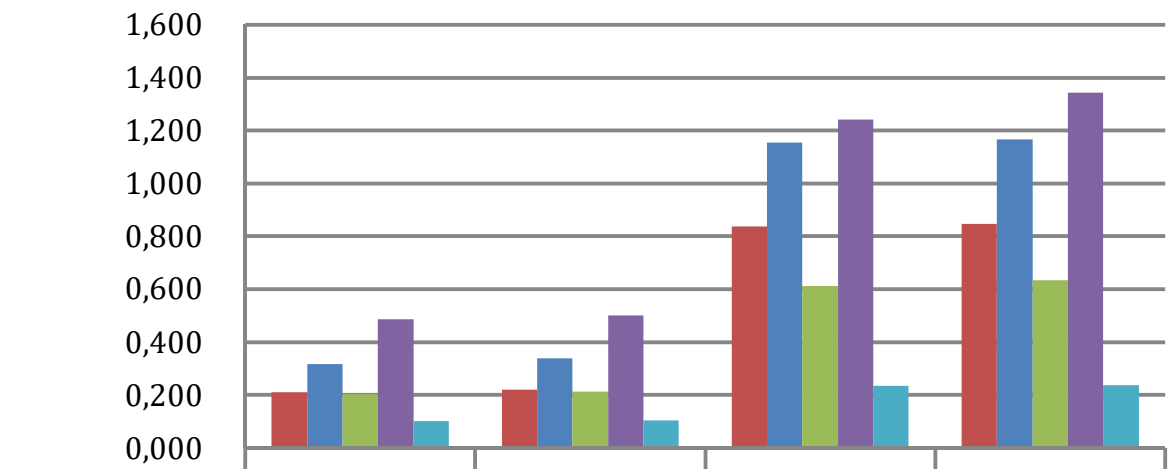

| | Grad-CAM | GE Grad-CAM | Excitation Backprop. | GE Excitation Backprop. |
|---|---|---|---|---|
| VGG-11 | 0,210 | 0,220 | 0,838 | 0,848 |
| VGG-16 | 0,317 | 0,339 | 1,156 | 1,166 |
| ResNet-50 | 0,202 | 0,214 | 0,612 | 0,635 |
| DenseNet-161 | 0,487 | 0,501 | 1,241 | 1,344 |
| MobileNet-v2 | 0,101 | 0,104 | 0,234 | 0,236 |

**Fig. 7.** Methods execution time in seconds—an average from 30 runs

As shown in the graph in Figure 7,the difference between the conventional method and the enhanced version is negligible, which is less than 10%. Note that because Gradual Extrapolation depends on an output from the original method its computation period would never be shorter than original procedure.

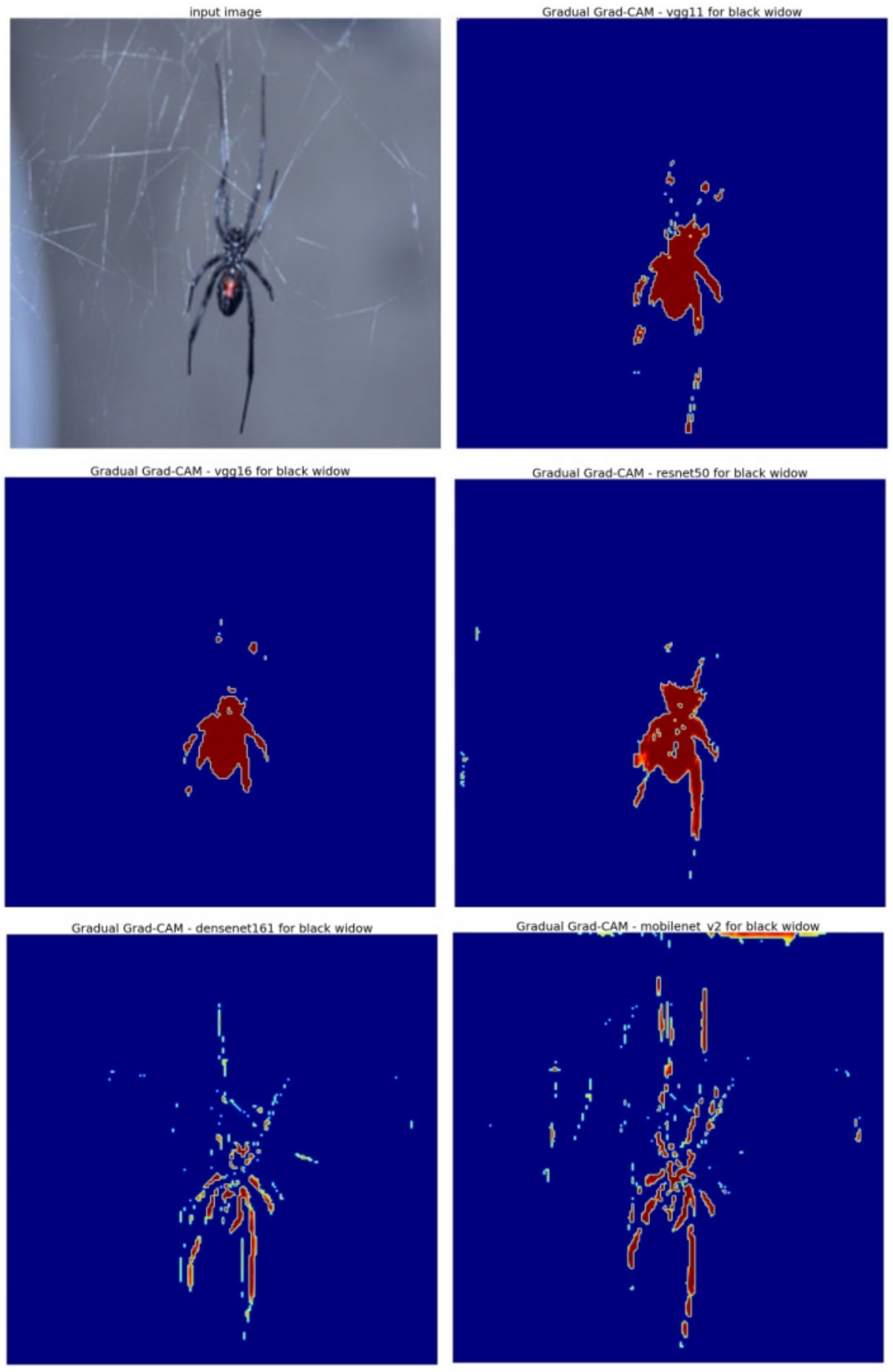

**Fig. 8.** Gradual Grad-CAM results of several models

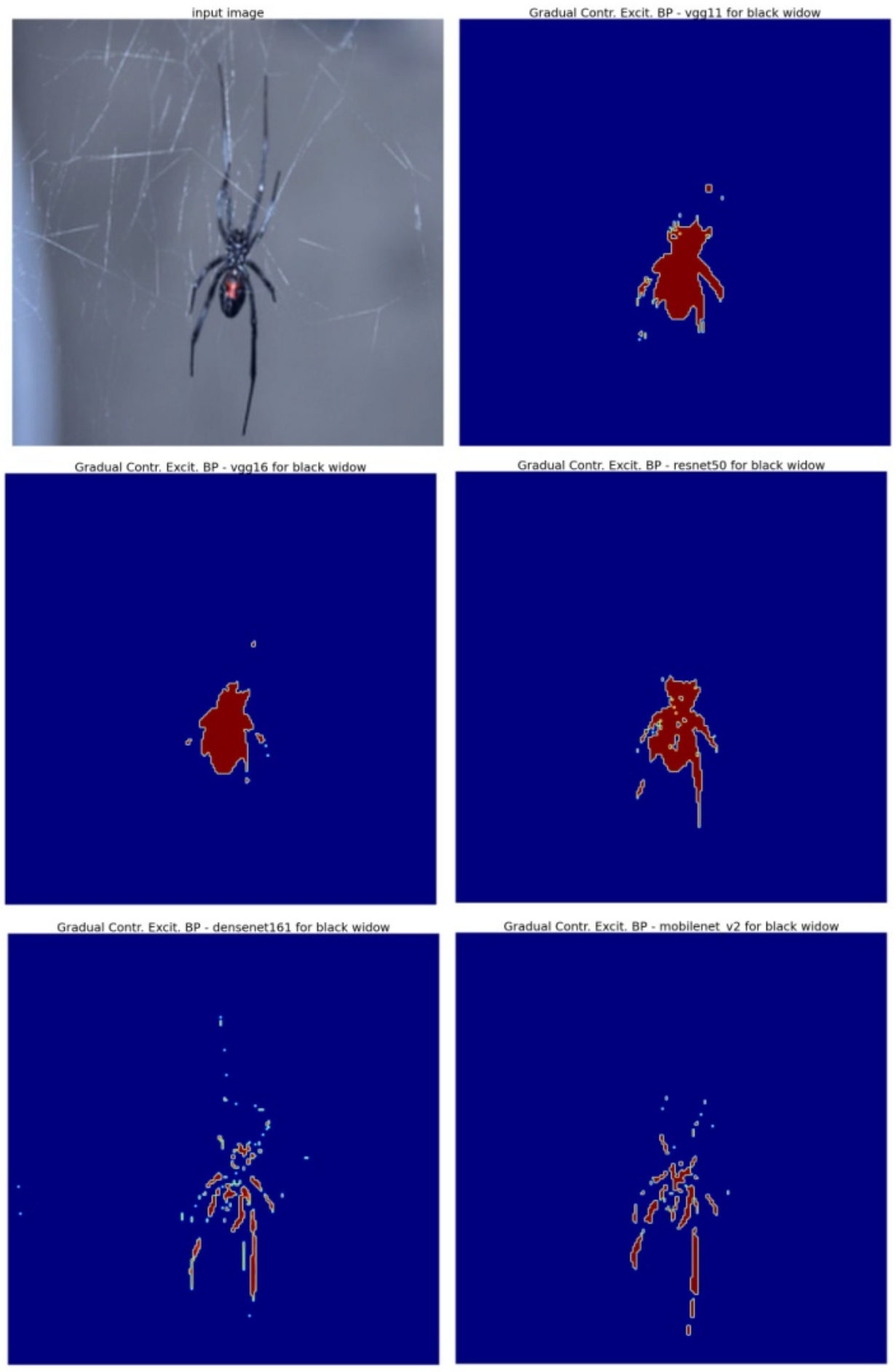

**Fig. 9.** Gradual contrastive excitation Backpropagation results of several models

The final experiment shows that Gradual Extrapolation is applicable to a wide variety of models. The following sample networks are used:

- VGG-11 [14]
- VGG-16 [14]
- ResNet50 [24]
- DenseNet-161 [25]
- MobileNet-V2 [26].

In each network, Gradual Extrapolated Grad-CAM and Contrastive Excitation Backpropagation are implementable. The results presented in Figures 8 and 9 prove the portability of the method to other models. Each image is identified by the network and can be used to localize the object and its shape.

## VI. CONCLUSIONS AND FUTURE WORKS

Herein, an enhancement technique is proposed that improves the explanations obtained using deep neural network methods. When combined with well-known methods such as Grad-CAM or Excitation Backpropagation, the proposed method improves the interpretability of the saliency map. Gradual Extrapolation often results in aesthetically sharper visualizations when applied to multilayer neural networks. It does not alter the attribution method itself, thus always inheriting the sensitivity of the underlying method. Moreover, it does not offer corrections if the original saliency is entirely incorrectly localized.

However, it may revise localization by highlighting the object shape, even if it extends beyond the original saliency region. Moreover, the proposed method reduces the background area accentuated by the saliency map.

Two subjectively selected heatmap generation techniques are analyzed. The proposed method is applicable to a wider range of visualization techniques that commonly refer to an original image via simple extrapolation.

To improve the proposed idea, experiments that obtain the mid layer importance matrices using more sophisticated methods are encouraged. Such methods are expected to further sharpen the resulting visualization.

The FIA tests prove the usefulness of the method as an enhancement of the original formula. The FIA tests confirm that Gradual Extrapolation often enhances visualizations by reducing information output and accuracy values at a very low cost with up to 10% longer computational time.

In summary, the method is not an explicit improvement of state-of-the-art saliency maps. It enhances their effectiveness in highlighting the shape of the object.

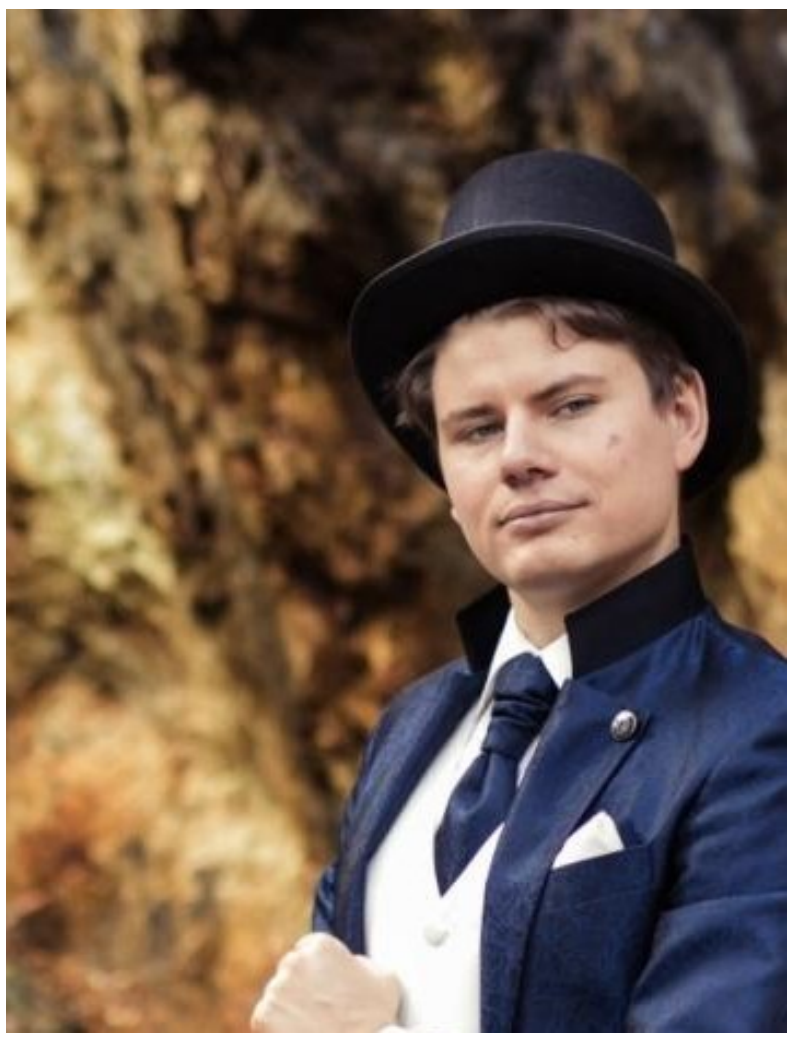

**TOMASZ SZANDAŁA** Assistant at the Wroclaw University of Science and Technology. Born in 1991 in Nysa, Poland. Finished his schooling in a catholic high school with distinction and started studying computer science at the Wroclaw University of Science and Technology. Simultaneously, he was working with IBM on Multipurpose Cloud Computing initiative, starting as a student and continuing as a senior mentor.Hisproclivity for solving computer science problems and sharing this knowledge with others sparked during membership in the student's scientific club: KREDEK Creation and Development Group. Obtained his master's degree along with the Dean's award for the top10 best graduates in 2015 in the specialty: Computer Science in Medicine.
After a year in the industry, he returned to the university to achieve a PhD in computer science. He is currently pursuing research work, which hasbeen published in numerous international publications, both scientific and technical.